\documentclass{article}
\usepackage{spconf}
\usepackage{amsmath,amssymb}
\usepackage{graphicx}
\usepackage{booktabs}
\usepackage{multirow}
\usepackage{tikz}
\usepackage{cite}
\usetikzlibrary{arrows.meta,positioning,fit,backgrounds,calc}
\usepackage[hidelinks]{hyperref}
\usepackage{orcidlink}

\tikzset{
  box/.style={draw, rounded corners=1.5pt, align=center, font=\scriptsize,
              minimum height=7mm, minimum width=14.5mm, inner sep=2pt},
  hl/.style={draw, rounded corners=1.5pt, align=center, font=\scriptsize,
              minimum height=7mm, minimum width=15mm, inner sep=2pt,
              fill=black!10, line width=0.8pt},
  ar/.style={-{Latex[length=1.5mm,width=1.2mm]}, semithick},
  lbl/.style={font=\tiny, inner sep=1pt}
}

\usepackage{microtype}
\let\origthebib\thebibliography
\def\thebibliography#1{\origthebib{#1}%
  \setlength{\itemsep}{0pt}\setlength{\parsep}{0pt}\setlength{\topsep}{2pt}}

\newcommand{\sd}[1]{{\tiny$\pm$#1}}
\newcommand{\fev}{\textit{few.val}}
\newcommand{\man}{\textit{many.val}}

\title{CAST-ASD: Causal Attention over Speakers and Time \\ for Online Active Speaker Detection}

\name{Pratham Gupta\,\orcidlink{0009-0000-7435-2074}$^{\,1,2,\dagger,\ast}$ \quad
      Haotian Qi\,\orcidlink{0009-0002-2946-0054}$^{\,2,\ast}$ \quad
      Tushar Sandhan\,\orcidlink{0009-0004-1138-4756}$^{\,1}$ \quad
      Gabriel Skantze\,\orcidlink{0000-0002-8579-1790}$^{\,2}$}

\address{
$^{1}$Department of Electrical Engineering, Indian Institute of Technology Kanpur \\
$^{2}$Department of Speech, Music and Hearing, KTH Royal Institute of Technology
}

\begin{document}
\ninept
\maketitle
{\renewcommand{\thefootnote}{}\footnotetext{$^{\dagger}$Corresponding author. \quad $^{\ast}$Equally contributing authors.}}

\begin{abstract}
We present CAST-ASD (Causal Attention over Speakers and Time), a causal joint model for active speaker detection (ASD). Recent ASD models improve performance by jointly processing all faces in a scene, but operate offline, with the full clip available before any frame is labelled. Low-latency causal detectors instead process each face independently. CAST-ASD combines joint face reasoning with fully causal inference. It applies scene-level attention across faces within the current frame and causal temporal attention over each face's history, enabling information sharing across speakers with zero lookahead. Its parameter count does not grow with the number of faces. On MSDWild, where most negative face-frames contain a visible competing speaker, CAST-ASD reduces competing-speaker false alarms by 32\% relative to the causal per-face detector it extends, with larger gains as scene occupancy increases. It outperforms all causal baselines on both validation splits while using only one-tenth the inter-speaker parameters of a three-block fixed-roster convolution. Code and models are available at \url{https://github.com/prathamg007/CAST-ASD}.
\end{abstract}


\section{Introduction}
\label{sec:intro}

Active speaker detection (ASD) determines, for every visible face in every
frame of a video, whether that person is speaking \cite{roth2020ava}. It is a
standard front end for audio-visual speaker diarization
\cite{liu2022msdwild,xu2022avaavd,medennikov2025sortformer}, for automatic
camera control in video conferencing \cite{cutler2020multimodal,gurvich2024rt},
for speech enhancement in hearing devices and augmented-reality glasses
\cite{donley2021easycom,huh2025egocentric}, and for
turn-taking models in spoken dialogue systems and social robots
\cite{skantze2021turn,ekstedt2022vap,qi2026muvap}. These systems operate online, often
under latency budgets of a few hundred milliseconds, so the detector that feeds
them must be causal. Most ASD models, and in particular the most accurate ones,
are not: they observe an entire clip before labelling any frame within it.

Detecting whether anyone is speaking is largely an acoustic question; deciding \emph{which} of several visible
faces is speaking is not, since all faces in a scene share one audio track and
therefore receive identical acoustic evidence. The discriminating evidence is
visual and is frequently weak at a single instant. Two sources of information
resolve such a frame: the future, obtained by waiting to see whose lips keep
moving, and the rest of the scene, obtained by comparing the candidates now.

Cross-face comparison, which we refer to as joint modelling, accounts for most
of the recent progress on AVA-ActiveSpeaker: context across co-occurring faces
\cite{alcazar2020asc,kopuklu2021asdnet}, graph formulations
\cite{leon2021maas,min2022spell}, unified context modelling
\cite{zhang2021unicon,alcazar2022easee}, the interleaved temporal attention and
fixed-roster convolution of LoCoNet \cite{wang2024loconet}, and the per-frame
speaker-interaction stream of D$^2$Stream \cite{xiao2025d2stream}. Models in
this line are trained and evaluated with bidirectional temporal modelling and
label a frame only after the clip containing it has been observed. Work aimed
at low latency has instead relied on lightweight per-face models
\cite{liao2023light,liao2025lrasd,huh2025egocentric}, on bounding a fusion
module to a fixed past and future window \cite{kundu2025streaming}, or on
spatial queries from a microphone array \cite{gurvich2024rt}, each of which
scores a face without access to the audio-visual evidence of its neighbours. A recent frame-level graph model
is described as compatible with streaming operation but does not report its
lookahead \cite{yin2025afsnet}. The motivation for avoiding joint architectures
is stated directly in \cite{kundu2025streaming}: their parameter counts are held
to be unsuitable for real-time use.

We show that joint modelling and causal inference are not in conflict. A comparison among
the faces present at one instant involves no time axis and is therefore causal
by construction, and we argue that it is worth more to a causal detector than to an offline
one because it supplies part of the evidence that lookahead would otherwise
have to provide.

\begin{table}[t]
\caption{Corpus statistics, computed with identical code over the validation
annotations of AVA-ActiveSpeaker and MSDWild. All entries are percentages of
face-frames. A negative face-frame is \emph{competing} when another tracked
face is genuinely speaking at that instant.}
\label{tab:corpus}
\vspace{0.6mm}
\centering
\setlength{\tabcolsep}{3.4pt}
\footnotesize
\begin{tabular*}{\columnwidth}{@{\extracolsep{\fill}}lccc@{}}
\toprule
 & & \multicolumn{2}{c}{MSDWild}\\
\cmidrule(l){3-4}
 & AVA & \fev & \man \\
\midrule
Scenes with $\geq 5$ faces       & \phantom{0}6.3 & \phantom{0}0.1 & 69.6 \\
Scenes with $\leq 3$ faces      & 88.8 & 80.0 & 15.4 \\
Competing negatives              & 19.4 & 79.7 & 76.2 \\
Overlapping speech               & \phantom{0}0.5 & \phantom{0}9.9 & 12.0 \\
\bottomrule
\end{tabular*}
\end{table}

Table~\ref{tab:corpus} contrasts AVA-ActiveSpeaker
with the MSDWild dataset~\cite{liu2022msdwild}, which contains in-the-wild conversational video
collected for multi-modal diarization. On AVA, a negative face-frame rarely
competes with another visible speaker and a three-face roster spans most of
the corpus, so systems above 95 mAP have been reported there
\cite{jung2024talknce} without having to pick a speaker from several candidates. Recent in-the-wild benchmarks report the
complementary finding: strong AVA accuracy does not carry over to
unconstrained multi-speaker video \cite{nguyen2025unitalk,roxo2024wasd}.
We therefore evaluate primarily on MSDWild, in which most negative face-frames
compete with a visible speaker. 

\begin{figure*}[t]
\centering
\includegraphics[width=\textwidth]{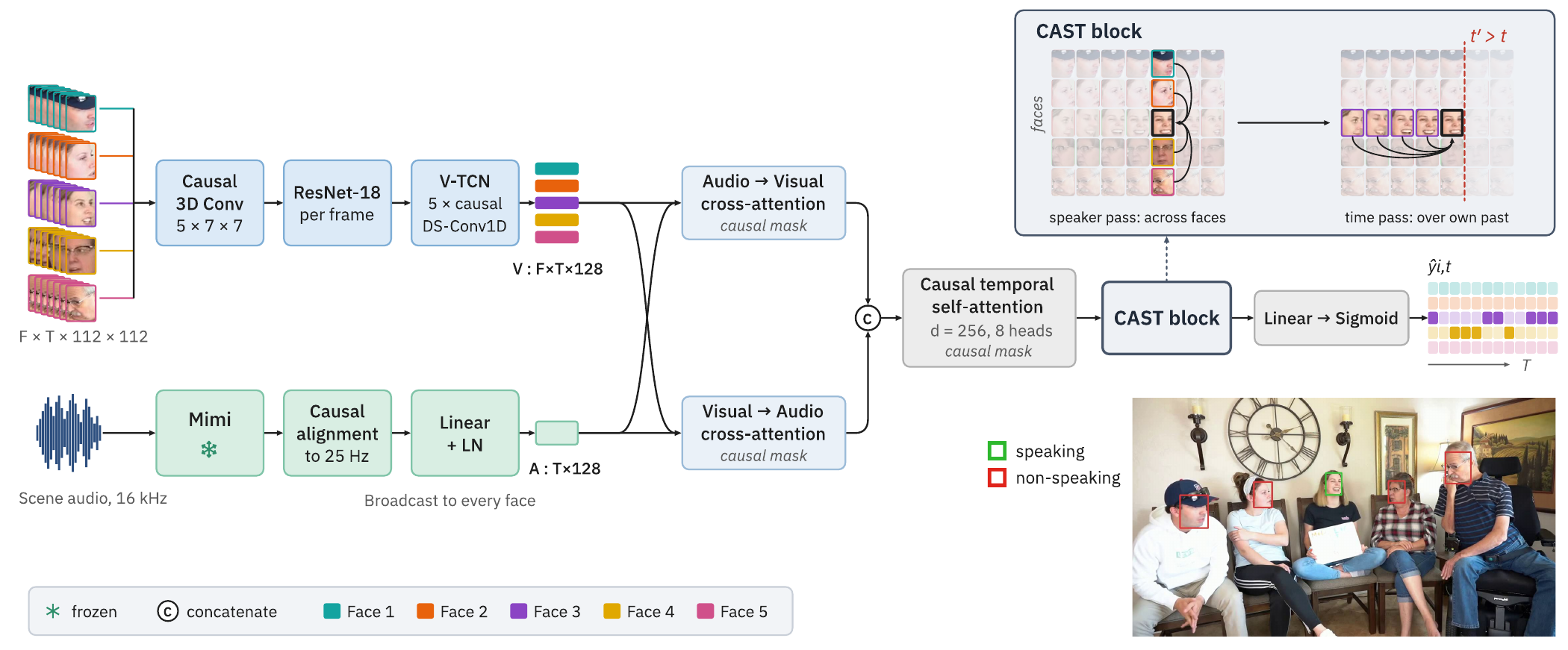}
\caption{CAST-ASD. The visual branch runs independently for each of the $F$
faces and the audio branch once per scene, so every face receives the same
mixed audio. The inset shows one CAST block: the speaker pass attends across
the faces present at a frame, the time pass attends over a face's own past
under a causal mask, and neither reads a future frame. Face crops are
grayscale, and are coloured here only to match each crop to its track. Right:
an example scene, with the active speaker boxed in green.}
\label{fig:arch}
\label{fig:block}
\end{figure*}

Our contributions are as follows.
1)~We propose CAST-ASD, a joint active speaker detector with zero lookahead. A
single CAST block, whose $1.57$M parameters do not depend on the number of
faces, lets each face attend to every other face at the current frame and to
its own past. On MSDWild it removes $32\%$ of the competing-speaker false alarms
of its per-face backbone and outperforms all causal baselines.
2)~We show that this improvement comes from the mixing operator, not the
audio front end. With the front end held fixed, a single CAST block yields
fewer competing-speaker false alarms than the three-block fixed-roster
convolution of LoCoNet while using one-tenth of its mixer parameters, and the
gap is largest in frames with more faces than the roster can hold.
3)~We measure the accuracy lost to causal inference for different
architectures trained under identical conditions on MSDWild; CAST-ASD loses
1.14 mAP, compared with $2.67$ mAP for LoCoNet.


\section{Method}
\label{sec:method}

\subsection{Problem formulation}
\label{sec:formulation}

A \emph{scene} is a set of $F$ face tracks that share one soundtrack. Given the
visual streams of the tracked faces and the single audio stream, an ASD model
produces a speaking probability $\hat{y}_{i,t}$ for every face $i$ and every
frame $t$. A \emph{joint} model allows each prediction to depend on all $F$ faces,
whereas a per-face model conditions the prediction for face $i$ on that face
and the audio alone. A model is \emph{causal} when perturbing any face or the audio at any
frame later than $t$ leaves $\hat{y}_{i,t}$ unchanged. These are the two
properties this work combines.

\subsection{Causal per-face backbone}
\label{sec:backbone}

CAST-ASD builds on a TalkNet-style backbone \cite{tao2021talknet} that scores
each face independently (Fig.~\ref{fig:arch}). Grayscale face crops pass
through a left-padded 3D convolution, a per-frame ResNet-18 \cite{he2016deep}
and five depthwise-separable causal 1D convolutions. 
The audio mixture is encoded using the frozen Mimi encoder from Moshi \cite{defossez2024moshi}. In the control experiments of Sec.~\ref{sec:operator}, we instead use LoCoNet's VGGish-based encoder \cite{hershey2017vggish,wang2024loconet}. In both cases, the audio features are causally aligned to the 25 Hz video frames.
Causally masked cross-attention is applied
in both directions between the two streams and the outputs are concatenated into a
$d\!=\!256$ embedding. A causal temporal self-attention layer then runs over
each face's fused stream, and the CAST block (Sec.~\ref{sec:cast}) is inserted
after it, before the head that produces one logit per face and frame. Audio is encoded once per scene and broadcast to
every face in it, which is what makes attribution a visual and contextual
problem.

Training minimises a binary cross-entropy loss on the audio-visual logit,
augmented with two auxiliary losses from heads that read the audio and visual streams
before fusion,
\begin{equation}
\mathcal{L} \;=\; \mathcal{L}_{av} \;+\; \lambda\,\mathcal{L}_{a}
\;+\; \lambda\,\mathcal{L}_{v},
\qquad \lambda = 0.4 ,
\label{eq:loss}
\end{equation}
averaged over the valid face-frames of the batch. The auxiliary terms keep
each modality independently predictive and are used unchanged for every system
compared here.

\subsection{The CAST block}
\label{sec:cast}

A scene is naturally a grid with one row per face and one column per frame, and
a CAST block reads along exactly one axis at a time (Fig.~\ref{fig:block}). Let
$H_{:,t}\in\mathbb{R}^{F\times d}$ denote the fused embeddings of all faces at
frame $t$, and $H_{i,:}\in\mathbb{R}^{T\times d}$ the trajectory of face $i$.
The \emph{speaker pass} compares the faces that are present at one frame using
multi-head self-attention \cite{vaswani2017attention},
\begin{equation}
Z_{:,t} \;=\; \operatorname{Attn}\!\left(H_{:,t}\right),
\qquad t = 1,\dots,T ,
\label{eq:speakerpass}
\end{equation}
and the \emph{time pass} then propagates the result along each face
individually under a mask $M$ that hides every frame later than the one being
computed,
\begin{equation}
U_{i,:} \;=\; H_{i,:} \,+\, \gamma\left(\operatorname{Attn}\!\left(Z_{i,:};\, M\right) - H_{i,:}\right),
\quad i = 1,\dots,F .
\label{eq:timepass}
\end{equation}
Here $\operatorname{Attn}$ is a Transformer layer \cite{vaswani2017attention} with eight-head attention followed by a feed-forward layer with expansion of
four, and $\gamma$ is a scalar gate on the block's output, initialised to zero, so an untrained block
reproduces the per-face backbone exactly. A block adds $1.57$M parameters and
the speaker pass costs $\mathcal{O}(F^2d)$ per frame.

Two properties follow. First, causality in the speaker pass is structural
rather than enforced by masking: it is evaluated on a tensor reshaped so that
faces occupy the sequence axis and no temporal axis is present, so only the
time pass requires a causal mask and the model remains causal for any number of
faces. Second, face slots carry no positional encoding, so the speaker pass is
permutation-equivariant \cite{lee2019set}: permuting the faces at a frame
permutes its outputs identically. This matters on conversational video, where
the ordering of face tracks is an arbitrary artefact of the tracker.

\subsection{Relation to fixed-roster convolution}
\label{sec:loconet}

The inter-speaker module of LoCoNet \cite{wang2024loconet} instead convolves
over the face axis, learning one weight matrix per pair of face slot and
temporal lag within a roster of $s$ faces, with $s\!=\!3$ in the
published model. Three consequences follow. Its reach across faces is fixed before training and cannot adapt to the scene,
so on \man{} a three-face roster spans only a small fraction of the faces
present (Table~\ref{tab:corpus}). Widening that reach is computationally expensive, since the
module learns $s$ weight matrices for a kernel of height $s$, so its parameter
count grows quadratically with the roster width. Unlike attention over an
unordered set, it is not permutation-equivariant, so its output depends on an
ordering that carries no information.

\section{Experimental Setup}
\label{sec:setup}

MSDWild \cite{liu2022msdwild} is segmented into
non-overlapping $30$\,s clips at $25$\,fps, the corpus provides two validation splits:
\fev{}, whose scenes hold two to four speakers, and \man{}, whose scenes hold
five or more. Training uses the few-speaker training split only, so all \man{}
results are zero-shot with respect to scene width. Reference speech whose speaker is not among the tracked faces accounts for
$3.5\%$ of \fev{} and $10.5\%$ of \man{}, bounding what any face-based
detector can recover. AVA-ActiveSpeaker \cite{roth2020ava} is derived from
Hollywood movies and provides face tracks of 1 to 10\,s, with $29{,}723$,
$8{,}015$ and $21{,}361$ tracks in its training, validation and test sets;
clips keep their native frame rate.

We report mAP over face-frames, computed with an
implementation that matches the official ActivityNet scorer to six decimal
places. mAP$_{\mathrm c}$ restricts the evaluation to frames in which another
visible face is genuinely speaking, and therefore measures attribution rather
than voice activity; FA$_{\mathrm c}$ is the false-alarm rate over those same
competing negatives, computed at a per-model threshold set to exactly $90\%$ recall
so that operating points are compared at matched recall. DER$^{\ast}$ is
diarization error with oracle tracking and oracle speaker identities, a
$0.25$\,s collar and overlap scored, which isolates the attribution component
of a diarization pipeline.

We compare against TalkNet \cite{tao2021talknet},
the ASD front end used in the MSDWild and AVA-AVD diarization pipelines
\cite{liu2022msdwild,xu2022avaavd}; Light-ASD \cite{liao2023light}, a
lightweight per-face model; and LoCoNet \cite{wang2024loconet}, a joint model
built on a fixed-roster inter-speaker convolution. Each baseline is retrained
with the audio front end released by its authors, MFCC features for TalkNet and
Light-ASD and a VGGish-based encoder \cite{hershey2017vggish} for LoCoNet, and
keeps its published classification loss. We also report our per-face backbone, the
causalized TalkNet-style backbone without scene-level mixing, using the frozen
Mimi encoder \cite{defossez2024moshi}.
Every architecture is trained in a causal and an offline
configuration that share the training protocol and seeds and differ only in the
causal constraints, namely the attention masks, the temporal padding of the
convolutional layers and the alignment of the audio encoder.

On MSDWild, all systems are trained for 12 epochs
with Adam (learning rate $2\times10^{-4}$, $5\%$ linear warm-up, cosine decay),
batches of up to $12{,}000$ padded face-frames and mixed precision, without
early stopping; the final epoch is reported. On AVA-ActiveSpeaker, all systems
use Adam at $10^{-4}$, decayed by $0.95$ after each epoch, for at most 25 epochs
with early stopping (patience 5) on validation mAP; every run stopped early.
Each configuration is trained with three seeds, one NVIDIA A100 GPU per run,
and we report the mean and seed standard deviation.

\section{Results}
\label{sec:results}

\subsection{Comparison of causal systems}
\label{sec:main}

\begin{table}[t]
\caption{Causal systems on the two MSDWild validation splits, with the offline
configuration of each architecture shown for reference. The audio encoder is
given in parentheses and $\times N$ denotes the number of mixer blocks. Bold
marks the best causal entry in each column.}
\label{tab:main}
\vspace{0.6mm}
\centering
\setlength{\tabcolsep}{2.6pt}
\scriptsize
\begin{tabular*}{\columnwidth}{@{\extracolsep{\fill}}lccccc@{}}
\toprule
& \multicolumn{4}{c}{\man{} ($5+$ speakers)} & \fev{}\\
\cmidrule(lr){2-5}\cmidrule(lr){6-6}
System & mAP & mAP$_{\mathrm c}$ & FA$_{\mathrm c}$ & DER$^{\ast}$ & mAP\\
\midrule
\multicolumn{6}{@{}l}{\emph{Causal: zero lookahead, verified by perturbation}}\\
\textbf{CAST-ASD} $\times1$ (Mimi) & \textbf{90.56}\sd{.10} & \textbf{71.51}\sd{.41} & \textbf{8.61}\sd{.18} & \textbf{27.59}\sd{.23} & \textbf{96.99}\sd{.02}\\
LoCoNet $\times3$ (VGGish) & 89.15\sd{.13} & 69.63\sd{.81} & 11.14\sd{.04} & 31.57\sd{.37} & 96.71\sd{.05}\\
Per-face backbone (Mimi) & 88.58\sd{.16} & 69.95\sd{.43} & 12.72\sd{.34} & 32.17\sd{.46} & 96.17\sd{.05}\\
TalkNet (MFCC) & 87.15\sd{.51} & 68.48\sd{.96} & 13.84\sd{.76} & 33.92\sd{.98} & 95.27\sd{.09}\\
Light-ASD (MFCC) & 85.10\sd{.22} & 65.78\sd{.42} & 16.73\sd{.35} & 38.16\sd{.21} & 94.22\sd{.08}\\
\midrule
\multicolumn{6}{@{}l}{\emph{Offline reference: reads the whole clip}}\\
CAST-ASD $\times1$ (Mimi) & 91.70\sd{.11} & 74.63\sd{.36} & 7.20\sd{.02} & 25.68\sd{.16} & 97.51\sd{.06}\\
CAST-ASD $\times3$ (Mimi) & 92.08\sd{.05} & 75.64\sd{.26} & 6.71\sd{.08} & 25.08\sd{.10} & 97.65\sd{.02}\\
LoCoNet $\times3$ (VGGish) & 91.82\sd{.34} & 75.35\sd{1.1} & 7.92\sd{.54} & 27.11\sd{.41} & 98.10\sd{.06}\\
TalkNet (MFCC) & 88.78\sd{.21} & 70.64\sd{.40} & 11.65\sd{.28} & 30.80\sd{.21} & 96.20\sd{.06}\\
Light-ASD (MFCC) & 88.64\sd{.28} & 71.03\sd{.81} & 12.18\sd{.55} & 32.13\sd{.71} & 95.96\sd{.13}\\
\bottomrule
\end{tabular*}
\end{table}

\begin{table}[t]
\caption{Mixing operator against audio front end on \man{}, three seeds per
row. The two mixers are exchanged between the two front ends at matched depth,
and the convolution is additionally given wider rosters  ($s\in\{4,5\}$) on our
own front end. Parameters are those of the scene mixer.}
\label{tab:operator}
\vspace{0.6mm}
\centering
\setlength{\tabcolsep}{3.6pt}
\renewcommand{\arraystretch}{1.05}
\scriptsize
\begin{tabular*}{\columnwidth}{@{\extracolsep{\fill}}llccrcc@{}}
\toprule
Audio & Mixer & $s$ & Blocks & Param. & mAP & FA$_{\mathrm c}$\\
\midrule
\multirow{6}{*}{Mimi}
  & \multirow{2}{*}{\textbf{CAST}} & --- & 1 & \phantom{0}1.57M & 90.56\sd{.10} & \phantom{0}8.61\sd{.18}\\
  &                                & --- & 3 & \phantom{0}4.72M & \textbf{90.65}\sd{.14} & \phantom{0}\textbf{8.44}\sd{.17}\\
\cmidrule(l){2-7}
  & \multirow{4}{*}{LoCoNet}       & 3 & 1 & \phantom{0}5.18M & 89.60\sd{.05} & 10.77\sd{.29}\\
  &                                & 3 & 3 & 15.55M & 89.91\sd{.24} & 10.26\sd{.42}\\
  &                                & 4 & 3 & 25.18M & 90.14\sd{.17} & \phantom{0}9.68\sd{.31}\\
  &                                & 5 & 3 & 37.57M & 90.37\sd{.15} & \phantom{0}9.42\sd{.18}\\
\midrule
\multirow{4}{*}{VGGish}
  & \multirow{2}{*}{\textbf{CAST}} & --- & 1 & \phantom{0}1.57M & 89.45\sd{.38} & \phantom{0}9.88\sd{.60}\\
  &                                & --- & 3 & \phantom{0}4.72M & 89.74\sd{.45} & \phantom{0}9.46\sd{.62}\\
\cmidrule(l){2-7}
  & \multirow{2}{*}{LoCoNet}       & 3 & 1 & \phantom{0}5.18M & 88.92\sd{.09} & 11.52\sd{.14}\\
  &                                & 3 & 3 & 15.55M & 89.15\sd{.13} & 11.14\sd{.04}\\
\bottomrule
\end{tabular*}
\end{table}

Table~\ref{tab:main} reports every causal system together with the offline
version of each architecture. CAST-ASD is the best causal system on both splits
and on every metric. The clearest result is on the metric the mechanism is designed to affect:
against the per-face backbone it extends, which shares its audio encoder,
visual encoder and classification head, the mixer lowers competing-speaker
false alarms on \man{} from $12.72$ to $8.61$ at matched recall, a relative
reduction of $32\%$. The advantage over LoCoNet in its published
configuration is also larger on mAP$_{\mathrm c}$, which scores only the
frames where attribution is at stake.

The effect carries through to the diarization metric, where CAST-ASD improves
DER$^{\ast}$ by $3.98$ points over LoCoNet (Table~\ref{tab:main}), and the causal model surpasses the
offline versions of TalkNet and Light-ASD, so a detector that reads no future
frames can outperform published architectures that read the whole clip.

\subsection{Mixing operator versus audio front end}
\label{sec:operator}

Each row of Table~\ref{tab:main} has different combinations of audio encoders and 
mixers, which supports conclusions about complete systems rather than individual components. Table~\ref{tab:operator} separates them by exchanging the front
ends at matched depth. The CAST block is ahead on both metrics in all four
matched pairs. The margin on FA$_{\mathrm c}$ is clear of the seed intervals
throughout, and a single CAST block has fewer competing false alarms than
three LoCoNet blocks on either front end while using one-tenth of their mixer
parameters.

The lower block of Table~\ref{tab:operator} gives the convolution its best
available configuration by widening the roster beyond the published value.
Both metrics improve with $s$, consistent with the operator's limited reach across faces, but that reach is bought at a parameter cost growing
quadratically with the roster: at $s\!=\!5$ the convolution still trails
CAST-ASD on both metrics while carrying 24 times as many mixer parameters.
Attention provides scene-wide reach without a width-dependent parameter cost
and with no roster width to fix in advance.

\subsection{Effect of scene width}
\label{sec:width}

\begin{table}[t]
\caption{Competing-speaker false alarms and mAP on \man{}, computed separately
by the number of faces visible at the frame (when more than one). Bold marks the
best causal entry in each column.}
\label{tab:occ}
\vspace{0.6mm}
\centering
\setlength{\tabcolsep}{3.4pt}
\scriptsize
\begin{tabular*}{\columnwidth}{@{\extracolsep{\fill}}lcccc@{}}
\toprule
Faces visible at the frame & 2 & 3 & 4 & $5+$\\
\midrule
Proportion of frames (\%) & 5.4 & 9.6 & 14.5 & 69.6\\
\midrule
\multicolumn{5}{@{}l}{\emph{Competing false alarms} FA$_{\mathrm c}$}\\
Per-face backbone (Mimi) & 8.25 & 9.97 & 17.17 & 12.45\\
LoCoNet $\times3$ (VGGish) & 6.94 & \textbf{7.02} & 13.33 & 11.39\\
\textbf{CAST-ASD} $\times1$ (Mimi) & 6.72 & 7.83 & 12.28 & \phantom{0}8.15\\
\textbf{CAST-ASD} $\times3$ (Mimi) & \textbf{6.30} & 7.15 & \textbf{11.56} & \phantom{0}\textbf{8.14}\\
\midrule
\multicolumn{5}{@{}l}{\emph{Average precision} mAP}\\
Per-face backbone (Mimi) & 96.74 & 93.66 & 83.99 & 86.87\\
LoCoNet $\times3$ (VGGish) & 97.01 & 94.13 & 85.59 & 87.28\\
\textbf{CAST-ASD} $\times1$ (Mimi) & \textbf{97.46} & \textbf{94.58} & 86.32 & 89.23\\
\textbf{CAST-ASD} $\times3$ (Mimi) & 97.35 & 94.50 & \textbf{86.79} & \textbf{89.32}\\
\bottomrule
\end{tabular*}
\end{table}

Table~\ref{tab:occ} reports results on \man{} depending on the number of faces visible at each
frame. CAST-ASD leads consistently on mAP, and its lead over
LoCoNet grows steadily as the number of faces rises. On FA$_{\mathrm c}$ the two joint
models are close while the scene fits inside LoCoNet's three-face roster and
separate sharply beyond it, the gap being largest for 5+ faces ($69.6\%$ of the split). This occupancy-dependent gap is the signature predicted by the proposed
mechanism and the most direct test of it: a convolution of fixed
kernel height cannot represent interactions it was not sized for.

\subsection{The cost of causal inference}
\label{sec:causalcost}

We measure the cost of causal inference as the difference in mAP between an architecture's offline and causal configurations, which, as described in Sec.~\ref{sec:setup}, use the same weight shapes, training recipe, and random seeds. Lookahead ablations have been reported for
individual models \cite{tao2021talknet,liao2025lrasd,kundu2025streaming}; what
they do not establish is how that cost varies across architectures evaluated
under identical conditions.

\begin{table}[t]
\caption{The cost of causal inference on \man{}, that is, the mAP an
architecture gives up when its causal constraints are applied.}
\label{tab:causal}
\vspace{0.6mm}
\centering
\setlength{\tabcolsep}{3pt}
\scriptsize
\begin{tabular*}{\columnwidth}{@{\extracolsep{\fill}}lccc@{}}
\toprule
Architecture & Causal & Offline & $\Delta$\\
\midrule
\textbf{CAST-ASD $\times1$ (Mimi)} & 90.56 & 91.70 & \textbf{1.14\sd{.11}} \\
CAST-ASD $\times3$ (Mimi) & 90.65 & 92.08 & 1.42\sd{.24}\\
TalkNet (MFCC) & 87.15 & 88.78 & 1.63\sd{.89}\\
LoCoNet $\times3$ (VGGish) & 89.15 & 91.82 & 2.67\sd{.59}\\
Light-ASD (MFCC) & 85.10 & 88.64 & 3.54\sd{.56}\\
\bottomrule
\end{tabular*}
\end{table}

Table~\ref{tab:causal} reports the comparison. CAST-ASD
suffers the smallest drop in mAP when causal constraints are applied, and does so from
the highest causal score in the table. CAST-ASD loses 1.14 mAP under causal constraints, whereas LoCoNet loses 2.67 mAP, more than twice as much.
This is consistent with lookahead and inter-face context acting as partial
substitutes, since either resolves which candidate owns the voice, and a mixer
whose reach is fixed in advance has less of the latter to draw on.

\subsection{Model capacity}
\label{sec:capacity}

Increasing the number of blocks does not improve accuracy. A single CAST block
already delivers the full gain (Table~\ref{tab:operator}): two and three blocks lie within seed variation
of one another, and six destabilise training, as they do for the convolutional
mixer. Since both mixers saturate at the same depth, we attribute this behavior to the task rather than to either architecture: identifying which visible face is speaking appears to require a single comparison, rather than multiple rounds of relational reasoning. Reversing the order of the two passes has no
measurable effect. One block is therefore the natural deployment
configuration, at $1.57$M parameters, $1.09$ GFLOP per face-frame and measured
zero lookahead.

\subsection{Generalisation to AVA-ActiveSpeaker}
\label{sec:ava}

\begin{table}[t]
\caption{AVA-ActiveSpeaker validation mAP, computed with the official
ActivityNet scorer. $^\dagger$Reported by the original authors under their own
training protocol.}
\label{tab:ava}
\vspace{0.6mm}
\centering
\setlength{\tabcolsep}{3.4pt}
\scriptsize
\begin{tabular*}{\columnwidth}{@{\extracolsep{\fill}}lcc@{}}
\toprule
System & Causal & Offline\\
\midrule
\textbf{CAST-ASD} $\times1$ (Mimi) & \textbf{91.07}\sd{.15} & 94.2\sd{.17}\\
LoCoNet $\times3$ (VGGish) & 90.19\sd{.06} & 95.2$^\dagger$\\
TalkNet (MFCC) & 89.11\sd{.18} & 92.3$^\dagger$\\
\bottomrule
\end{tabular*}
\end{table}

Table~\ref{tab:ava} repeats the comparison on
AVA-ActiveSpeaker under a single training recipe, so the systems differ
only in audio encoder and scene mixer. CAST-ASD is ahead of both, with one
block against LoCoNet's three. The ordering is thus reproduced on a corpus
whose statistics, by Table~\ref{tab:corpus}, lie close to the regime in which
a three-face roster suffices, and the margin is correspondingly smaller than
on MSDWild, as the mechanism predicts.

\section{Conclusion}
\label{sec:conclusion}

Joint modelling and causal inference have been treated as alternatives in
active speaker detection, one line of work buying accuracy by reading the
whole clip and the other buying latency by scoring faces in isolation. We have
shown that they are compatible. Attention across the faces present at one
instant introduces no time axis, adds $1.57$M parameters per block whatever
the scene contains, and removes a third of the competing-speaker false alarms
of the causal per-face detector it extends. CAST-ASD is the strongest causal
system in our comparison on both MSDWild splits and on AVA-ActiveSpeaker. It is ahead of a fixed-roster convolution with a matched front end at one-tenth of the inter-speaker parameters, and it lowers the cost of causal inference from
$2.67$ to $1.14$ mAP. Its advantage grows with the number of faces on screen,
the regime online diarization and multiparty dialogue systems operate in.
Future work will route probability mass explicitly when the speaker is off
camera, treat overlapped speech jointly \cite{makishima2025which}, add
identity cues \cite{clarke2025speaker}, and evaluate on newer corpora \cite{nguyen2025unitalk,roxo2024wasd} against
recent models \cite{wang2026gatefusion}.

\section{Acknowledgment}

This work was supported by the Wallenberg AI, Autonomous Systems and Software Program (WASP), funded by the Knut and Alice Wallenberg Foundation (KAW), and the Swedish Research Council project 2020-03812. Computations and data handling were enabled by the Berzelius resource at the National Supercomputer Centre, provided by the Knut and Alice Wallenberg Foundation. The authors declare no conflicts of interest.

\noindent\textbf{Compliance with Ethical Standards:}
This work uses human-subject data released in open access by
AVA-ActiveSpeaker~\cite{roth2020ava} and MSDWild~\cite{liu2022msdwild};
ethical approval was not required under their licences.

\bibliographystyle{IEEEbib}
\bibliography{refs}

\end{document}